\documentclass[letterpaper, 10 pt, conference]{ieeeconf}  

\IEEEoverridecommandlockouts                              

\usepackage{amsmath}
\usepackage{bm}
\usepackage{amsfonts}
\usepackage{multirow}
\usepackage{graphicx}
\usepackage{subfig}
\usepackage{tikz}
\usepackage{cite}
\usepackage{subfig}
\usepackage{xcolor}
\usepackage{graphicx}
\usepackage{mwe}
\let\labelindent\relax 
\usepackage{enumitem}
\usepackage{adjustbox}
\usepackage[most]{tcolorbox}

\usepackage{algorithm2e}
\usepackage[font=footnotesize]{caption}
\usepackage[noend]{algpseudocode}
\usepackage{hyperref,cleveref}
\usepackage[colorinlistoftodos,prependcaption,textsize=tiny]{todonotes}
\usepackage{acronym}
\usepackage{pgfplots}
\pgfplotsset{compat=1.15}
\newlength\Fcolumnseprule
\usepackage[colorinlistoftodos]{todonotes}

\acrodef{CFAR}		[CFAR]			{Constant False-Alarm Rate}
\acrodef{FMCW}		[FMCW]			{Frequency-Modulated Continuous Wave}
\acrodef{TITLE}		[CFEAR Radarodometry]			{Conservative Filtering for Efficient and Accurate Radar Odometry}

\title{\LARGE \bf

Oriented surface points for efficient and accurate radar odometry
}

\author{Daniel Adolfsson, Martin Magnusson, Anas Alhashimi, Achim J. Lilienthal, Henrik Andreasson 
  \thanks{The authors are with the MRO lab of the AASS research centre
    at \"Orebro University, Sweden.
    E-mail: \texttt{Daniel.Adolfsson@oru.se}}%
  \thanks{This work has received funding from the Swedish Knowledge Foundation (KKS) project ``Semantic Robots'' and European Union's
    Horizon~2020 research and innovation programme under grant
    agreement No 732737 (ILIAD).}%
}

\begin{document}

\maketitle
\thispagestyle{empty}
\pagestyle{empty}

\begin{abstract}

This paper presents an efficient and accurate radar odometry pipeline for large-scale localization. We propose a radar filter that keeps only the strongest reflections per-azimuth that exceeds the expected noise level. The filtered radar data is used to incrementally estimate odometry by registering the current scan with a nearby keyframe.
By modeling local surfaces, we were able to register scans by minimizing a point-to-line metric and accurately estimate odometry from sparse point sets, hence improving efficiency.
Specifically, we found that a point-to-line metric yields significant improvements compared to a point-to-point metric when matching sparse sets of surface points. 
Preliminary results from an urban odometry benchmark show that our odometry pipeline is accurate and efficient compared to existing methods with an overall translation error of 2.05\%, down from 2.78\% from the previously best published method, running at 12.5~ms per frame without need of environmental specific training.


\end{abstract}

\section{Introduction}
Radars are promising sensors for robust perception in robotics and autonomous navigation. The sensor can operate in harsh environments and extreme weather due to its ability to penetrate snow, rain, fog and dust. However, radar data is notoriously hard to interpret due to sensing artifacts such as speckle noise, multi-path reflections and saturation. In order to overcome sensing artifacts for the task of estimating large-scale radar odometry, authors have proposed various techniques such as dense matching~\cite{9197231,barnes_masking_2020}, image feature extraction~\cite{hong2020radarslam}, landmark detection~\cite{8460687}, learning to predict key points~\cite{barnes_under_2020} or mask noise~\cite{barnes_masking_2020}. Sparse methods mitigate false detections during data association by finding the largest common subset ~\cite{8460687,hong2020radarslam} or via dense search~\cite{barnes_under_2020}. Existing learning methods requires an extensive amount of data~\cite{barnes_masking_2020,barnes_under_2020} and a ground truth positioning system to supervise the learning, despite this they provide limited generalization. Dense methods are inefficient and do not scale well with spatial resolution. Methods using landmarks or image features based  have achieved limited accuracy, despite using robust data association algorithms to mitigate the effect of outliers and sensing artifacts~\cite{8460687,hong2020radarslam}.
In this work we propose an efficient and accurate pipeline for radar odometry estimation. Radar data is filtered per azimuth keeping the strongest returns that exceed the expected noise level. The filtered data is used to compute a small number of oriented surface points that efficiently represent the local geometry of the environment. Sets of oriented surface points are registered by iteratively by associating pairs and minimizing a point-to-line metric. Preliminary results indicate that our method is both efficient and accurate compared to previous methods with the additional benefit of being \emph{learning-free} and comfortably runs on a single CPU core without requiring training data.

\section{Method}
\begin{figure}
\vspace{0.2cm}
    \centering
\includegraphics[width=0.99\hsize,trim={0cm 0cm 2cm 0cm},clip]{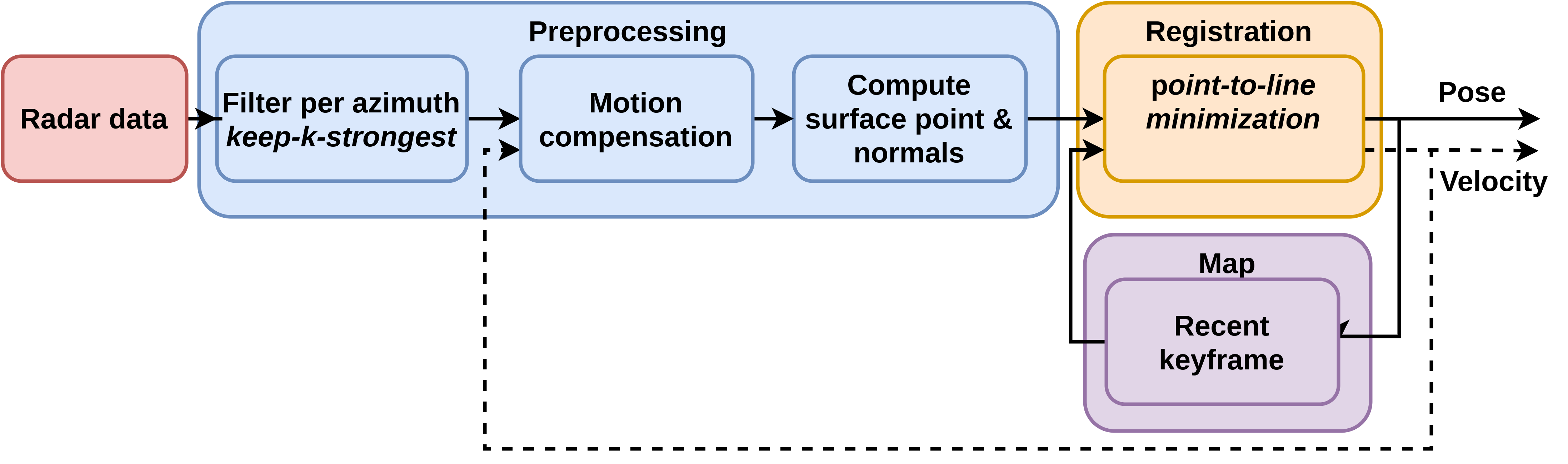}
    \caption{Overview of radar odometry pipeline.}
    \label{fig:overview}
    
\end{figure}

An overview of the method is depicted in Fig.~\ref{fig:overview}.
\paragraph{Prepossessing} Our pipeline operates on a rotating 2d radar that provides data $\mathbf{Z}_{m \times n}$ on polar form with $m$ azimuths and $n$ range bins per azimuth.
For each azimuth (row), the $k=12$ strongest reflections (columns), exceeding a minimum level $z_{min}$ are kept and converted to 2D Cartesian space.
A cell grid with resolution $(r/2)$ is used to discretize space. Around each cell center with nearby points, we compute an oriented surface point (sample mean and normal $\{\bm{\mu}_i,\bm{n}_i\}$) from all points within a radius $r$. The normal is computed from the smallest eigenvector of the sample covariance of nearby points. All oriented surface points computed from sweep $t$ are added to $\bm{\mathcal{M}}^t$.

We compensate for motion distortion using the previous velocity estimate $\bm{v}_{t-1}$, assuming acceleration is small.

\paragraph{Registration}
Each oriented surface point $\{\bm{\mu}_i,\bm{n}_i\}\in\bm{\mathcal{M}}^t$ is assigned one correspondence by searching for neighbours within a radius $r$ in the nearest keyframe $\bm{\mathcal{M}}^k$. The relative alignment $\bm{x}$ between $\bm{\mathcal{M}}^t$ and recent keyframe $\bm{\mathcal{M}}^k$ is then found by solving 
\begin{equation}
    \underset{\mathbf{x}}{\mathrm{arg\: min}} f(\bm{\mathcal{M}}^{k},\bm{\mathcal{M}}^t,\bm{x}),
\end{equation}
where $f$ computes the sum of point-to-line (P2L) distances for each pair of correspondences. The velocity $\bm{v}_t$ is approximated from the last two pose estimates $\bm{v}_t=\frac{1}{\Delta t}(\bm{x}_t-\bm{x}_{t-1})$. When the distance between current and keyframe position exceeds a threshold, a new key-frame is created at $\bm{x}_t$.

\section{Evaluation}

\begin{table*}
\centering
 \begin{adjustbox}{width=\textwidth}

\begin{tabular}{l|ll|lllllllll|l|l}
              & & & \multicolumn{1}{l}{\textbf{Sequence}}    &                                                                        \\
\textbf{Method} & \textbf{Evaluation} & \textbf{resolution}        & 10-12-32 & 16-13-09 & 17-13-26 & 18-14-14 & 18-15-20 & 10-11-46 & 16-11-53 & 18-14-46 & Mean  & \textbf{Mean SCV}   \\
\hline \\
Visual Odometry  ~\cite{Churchill2012ExperienceBN} & ~\cite{barnes_masking_2020} & N/A   & N/A        & N/A        & N/A        & N/A        & N/A        & N/A        & N/A        & N/A        &    $3.78/0.01$ & N/A        \\
\hline \\
SuMa (Lidar)~\cite{behley2018rss}  & ~\cite{hong2020radarslam }& N/A  & $1.1/0.3$ & $1.2$/$0.4$ & $1.1/0.3$ & $0.9/0.1$ & $1.0/0.2$  & $1.1/0.3$        & $0.9/0.3$        & $1.0/0.1$        & $1.16/0.3$    & $1.03/0.3$        \\
\hline \\
Cen~\cite{8460687}    &   ~\cite{barnes_masking_2020} & $0.175$& N/A        & N/A        & N/A        & N/A        & N/A        & N/A        & N/A        & N/A        & $3.72/0.95$    & $3.63/0.96$       \\
Robust Keypoints~\cite{barnes_under_2020}    &  & $0.346$    & N/A        & N/A        & N/A        & N/A        & N/A        & N/A        & N/A        & N/A        & $2.05/0.67^*$    & N/A  &         \\
Barnes Dual Cart~\cite{barnes_masking_2020} & & $0.043$ & N/A        & N/A        & N/A        & N/A        & N/A        & N/A        & N/A        & N/A        & $\bm{1.16/0.3}^*$    &   $2.784/0.85 $ \\
Hong odometry~\cite{hong2020radarslam} & &$0.043$ & $2.98/0.8$        & $3.12/0.9$        & $2.92/0.8$        & $3.18/0.9$        & $2.85/0.9$        & $3.26/0.9$        & $3.28/0.9$        & $3.33/1$        & $3.11/0.9$    & $3.11/0.9$         \\
 (ours) &  & $0.043$ & \textbf{1.81/0.58/0.057}          & \textbf{2.23/0.68/0.067}        & \textbf{2.04/0.66/0.057}        & \textbf{1.894/0.61/0.063}        & \textbf{1.83/0.52/0.078}        & \textbf{2.15/0.67/0.063}        & \textbf{2.38/0.68/0.07}        & \textbf{2/0.62/0.063}        & 2.05/0.63/0.066
    & \textbf{2.05/0.63/0.066}

\end{tabular}
\end{adjustbox}

\caption{Comparison of odometry methods by their translation error [\%] and rotation error [deg/100m]). For our method we include the relative pose error as a the last value per column. \cite{barnes_masking_2020,barnes_under_2020} should be compared based on their Mean Spatial Cross  Validation (SCV) error to make the evaluation fair.
}
\label{tab:results}
  
\end{table*}

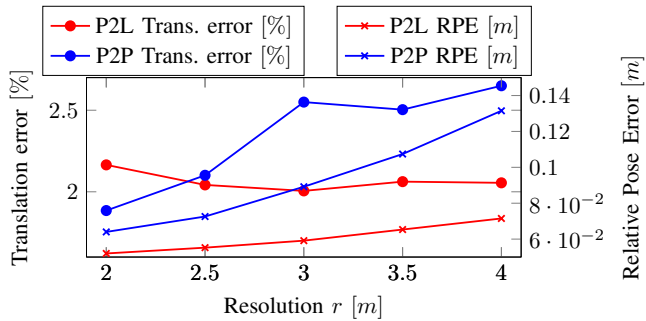
\begin{figure}
    \centering
    \begin{adjustbox}{width=\hsize}
\begin{tikzpicture}
	\begin{axis}
	[
		width   = 0.95\hsize,
		height  = 0.5\hsize,
		xmin		= 1.9,
		xmax		= 4.1,
		ymin		= 1.6,
		ymax		= 2.7,
		xlabel  = {Resolution $r$ $[m]$},
		ylabel  = {Translation error $[\%]$},
		ylabel near ticks,
		legend style={at={(0.2,1.4)},anchor=north},
		legend columns = 1,
	]

	\addplot
	[
		white!25!black,
		thick,
		solid,
		mark=*,
		red,
	]
	table
	[
		x   = Resolution,
		y   = Translation_error,
	]
	{Datasets/comparsion_p2l_data.txt};
    	\addlegendentry{P2L Trans. error $[\%]$ };

	\addplot
	[
		white!25!black,
		thick,
		solid,
		mark=*,
		blue,
	]
	table
	[
		x   = Resolution,
		y   = Translation_error,
	]
    {Datasets/comparison_p2p_data.txt};
    \addlegendentry{P2P Trans. error $[\%]$ };
    

	\end{axis}
	\begin{axis}
	[
		width   = 0.95\hsize,
		height  = 0.5\hsize,
		xmin		= 1.9,
		xmax		= 4.1,
		ymin		= 0.05,
		ymax		= 0.15,
		ylabel  = {Relative Pose Error $[m]$},
		axis y line*=right,
		ylabel near ticks,
		legend style={at={(0.8,1.4)},anchor= north},
		legend columns = 1,
	]
		\addplot
	[
		white!25!black,
		thick,
		solid,
		mark=x,
		red,
	]
	table
	[
		x   = Resolution,
		y   = RPE_m,
	]
    {Datasets/comparsion_P2L_data.txt};
    \addlegendentry{P2L RPE $[m]$};

	\addplot
	[
		white!25!black,
		thick,
		solid,
		mark=x,
		blue,
	]
	table
	[
		x   = Resolution,
		y   = RPE_m,
	]
    {Datasets/comparison_p2p_data.txt};
    \addlegendentry{P2P RPE $[m]$};


	\end{axis}

\end{tikzpicture}
\end{adjustbox}
    \vspace{-0.2cm}
    \caption{odometry accuracy in the sequences 18-14-14 and 16-13-09. Each point is computed from odometry error averaged over $20$~km. P2P and P2L are compared by their translation error and Relative Pose Error (RPE).
    A courser resolution increase the translation error and RPE, especially when matching P2P.}
    \label{fig:comparison_p2p_p2l}
    \vspace{0cm}
\end{figure}

We evaluated our method using the Oxford Radar RobotCar Dataset and follow the standard odometry benchmark proposed in \cite{Geiger2012CVPR} to compute odometry performance metrics.
To understand the benefit of considering the local geometry in the cost function, we compared point-to-line (P2L) to a point-to-point (P2P) metric by varying the resolution over two sequences in the Oxford dataset, the result is depicted in Fig.~\ref{fig:comparison_p2p_p2l}. Using P2L makes the method more accurate compared to P2P, especially when using a coarser grid (larger radius $r$) which makes the point set sparser.
We also compared our pipeline with other methods that have been previously evaluated on the Oxford Radar dataset. Learning-based methods are evaluated based on their Spatial Cross Validation (SCV).
The result is presented in \cref{tab:results}. Our method scores between 1.81-2.38\% translation error with an average of 2.05\%. This is more accurate than comparable methods such as Hong (3.11\%), Cen (3.63\%), and Mean SCV Barnes Dual Cart (2.784\%).

The overall computation time per frame was $12.5\pm3.6$~ms, $4.8\pm 1.4$~ms for filtering and $7.7\pm 2.2$~ms for computing and iteratively match surface points.

\begin{figure}
    \centering
    \includegraphics[width=0.6\hsize,trim={10cm 1.5cm 10cm 1.5cm},clip]{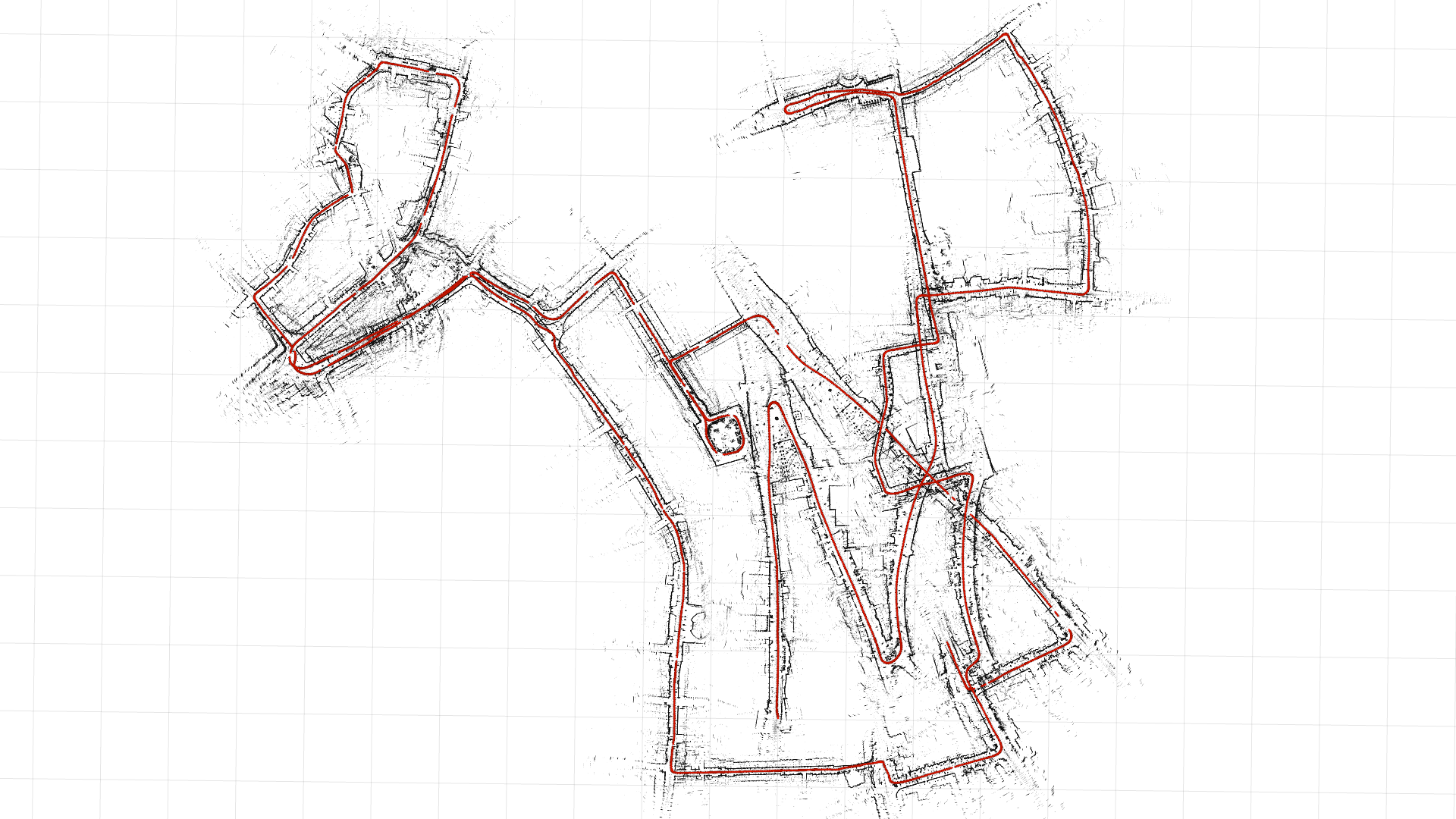}
    \caption{Estimated odometry in sequence 12-10-32 of the Oxford dataset.}
    \label{fig:oxford_odometry}
    
\end{figure}

\section{Discussion and future work}

Filtering originally noisy radar data to a small and clean set of points allows us to compute a sparse set of oriented surface points. Considering the normals in the registration cost function makes the method less sensitive to sparsity, and hence enables efficient and accurate matching.
We believe that considering the local geometry is important when matching sparse set of points, especially via one-to-one correspondence (one neighbor per surface point).
The difference between P2L and P2P seems to be lower for denser point clouds (lower resolution) as shown by the evaluation. This is intuitive as a denser point cloud inherently provides more information about the underlying geometry without explicitly modeling surfaces.

Surprisingly, a relative high odometry accuracy was achieved using a simple heuristic that filters radar data based on the $k$ strongest returns exceeding the expected noise level $z_{min}$. Limiting the maximum returns to e.g. $k=12$ per azimuth can be too conservative for some applications and potentially filter important landmarks. However, the limitation also makes the filter largely insensitive to $z_{min}$ and $k$, and the noise level $z_{min}$ can be chosen fairly low without introducing an excessive amount of false detections. For that reason, we hypothesize that the filter can generalize to new environments without needing to change the parameters, and hope that the presented odometry pipeline can serve as a basis to a versatile, flexible and highly robust localization system.







\addtolength{\textheight}{-5cm}   
\bibliographystyle{IEEEtran}
\bibliography{Conference/references.bib}

\begin{thebibliography}{1}
\providecommand{\url}[1]{#1}
\csname url@rmstyle\endcsname
\providecommand{\newblock}{\relax}
\providecommand{\bibinfo}[2]{#2}
\providecommand\BIBentrySTDinterwordspacing{\spaceskip=0pt\relax}
\providecommand\BIBentryALTinterwordstretchfactor{4}
\providecommand\BIBentryALTinterwordspacing{\spaceskip=\fontdimen2\font plus
\BIBentryALTinterwordstretchfactor\fontdimen3\font minus
  \fontdimen4\font\relax}
\providecommand\BIBforeignlanguage[2]{{%
\expandafter\ifx\csname l@#1\endcsname\relax
\typeout{** WARNING: IEEEtran.bst: No hyphenation pattern has been}%
\typeout{** loaded for the language `#1'. Using the pattern for}%
\typeout{** the default language instead.}%
\else
\language=\csname l@#1\endcsname
\fi
#2}}

\bibitem{barnes_under_2020}
D.~{Barnes} and I.~{Posner}, ``Under the radar: Learning to predict robust
  keypoints for odometry estimation and metric localisation in radar,'' in
  \emph{(ICRA)}, 2020, pp. 9484--9490.

\bibitem{barnes_masking_2020}
D.~Barnes, R.~Weston, and I.~Posner, ``Masking by moving: Learning
  distraction-free radar odometry from pose information,'' in \emph{CoRL}, ser.
  CoRL, L.~P. Kaelbling, D.~Kragic, and K.~Sugiura, Eds., vol. 100.\hskip 1em
  plus 0.5em minus 0.4em\relax PMLR, 30 Oct--01 Nov 2020, pp. 303--316.

\bibitem{behley2018rss}
J.~Behley and C.~Stachniss, ``Efficient surfel-based {SLAM} using {3D} laser
  range data in urban environments,'' in \emph{
  Systems~ (RSS)}, 2018.

\bibitem{8460687}
S.~H. {Cen} and P.~{Newman}, ``Precise ego-motion estimation with
  millimeter-wave radar under diverse and challenging conditions,'' in
  \emph{2018 IEEEn (ICRA)}, 2018, pp. 6045--6052.

\bibitem{Churchill2012ExperienceBN}
W.~S. Churchill, ``Experience based navigation : theory, practice and
  implementation,'' 2012.

\bibitem{Geiger2012CVPR}
A.~Geiger, P.~Lenz, and R.~Urtasun, ``Are we ready for autonomous driving? the
  {KITTI} vision benchmark suite,'' in \emph{
  Pattern Recognition (CVPR)}, 2012.

\bibitem{hong2020radarslam}
Z.~Hong, Y.~Petillot, and S.~Wang, ``Radarslam: Radar based large-scale slam in
  all weathers,'' in \emph{2020 (IROS)}, 2020, pp. 5164--5170.

\bibitem{9197231}
Y.~S. {Park}, Y.~S. {Shin}, and A.~{Kim}, ``Pharao: Direct radar odometry using
  phase correlation,'' in \emph{2020 IEEE (ICRA)}, 2020, pp. 2617--2623.

\end{thebibliography}

\end{document}